\documentclass[11pt,a4paper]{article}
\usepackage{pgfplots}
\usepackage{authblk}
\usepackage[hyperref]{emnlp2018}
\usepackage{times}
\usepackage{latexsym}
\usepackage{url}
\usepackage{amsmath}
\usepackage{amssymb}
\usepackage{graphicx}
\usepackage{caption}
\usepackage{subcaption}
\usepackage{alltt}
\usepackage{ctable}
\usepackage{multirow}
\usepackage{dsfont}
\usepackage{placeins}
\usepackage{xcolor,colortbl}
\usepackage{xifthen}
\usepackage{float}
\usepackage{mathtools}

\aclfinalcopy % Uncomment this line for the final submission
\colorlet{CellColor}{blue!50}
\colorlet{CellColorCCO}{blue!100}
\colorlet{CellColorCCC}{purple!100}

\def\todo#1{\bgroup \textcolor{red}{(#1)}\egroup}

\DeclareMathOperator*{\argmin}{\arg\min}  
\DeclareMathOperator*{\argmax}{\arg\max}

\title{Linear Transformations for Cross-lingual Semantic Textual Similarity}

\author[ ]{\bf Tom\'{a}\v{s} Brychc\'{i}n}
\affil[ ]{NTIS -- New Technologies for the Information Society,}
\affil[ ]{Faculty of Applied Sciences, University of West Bohemia, Czech Republic}
\affil[  ]{\tt	brychcin@kiv.zcu.cz}
\affil[  ]{\tt	http://nlp.kiv.zcu.cz}

\begin{document}
\maketitle
\begin{abstract}
Cross-lingual semantic textual similarity systems estimate the degree of the meaning similarity between two sentences, each in a different language.
State-of-the-art algorithms usually employ machine translation and combine vast amount of features, making the approach strongly supervised, resource rich, and difficult to use for poorly-resourced languages.

In this paper, we study linear transformations, which project monolingual semantic spaces into a shared space using bilingual dictionaries.
We propose a novel transformation, which builds on the best ideas from prior works. 
We experiment with unsupervised techniques for sentence similarity based only on semantic spaces and we show they can be significantly improved by the word weighting.
Our transformation outperforms other methods and together with word weighting leads to very promising results on several datasets in different languages.
\end{abstract}

\section{Introduction}
%\vspace{-0.2cm}

Semantic textual similarity (STS) systems estimate the degree to which two textual fragments (e.g., sentences) are semantically similar to each other.
STS systems are usually evaluated by human judgments. The ability to compare two sentences in meaning is one of the core parts of natural language understanding (NLU), with applications ranging across machine translation, summarization, question answering, etc. 

%SemEval (International Workshop on Semantic Evaluation) encourages and supports research in this area and since 2012 it has held the STS shared tasks annually, allowing independent evaluation of state-of-the-art algorithms.
SemEval (International Workshop on Semantic Evaluation) has held the STS shared tasks annually since 2012.
During this time, many different datasets and methods have been proposed.
Early methods focused mainly on surface form of sentences and employed various word matching algorithms \cite{bar-EtAl:2012:STARSEM-SEMEVAL}.
\newcite{han-EtAl:2013:SEM} added distributional word representations and WordNet, achieving the best performance at SemEval 2013. 
Word-alignment methods introduced by \newcite{sultan:2015} yielded the best correlations at SemEval 2014 and 2015.
Nowadays, the best performance tends to be obtained by careful feature engineering combining the best approaches from previous years together with deep learning models \cite{rychalska-EtAl:2016:SemEval,tian-EtAl:2017:SemEval}.

% On top of that, open and robust MT
%models still do not exist for many language pairs, which further impedes the
%wide applicability of the proposed CL STS models. Most of the monolingual
%STS models (most of which are for English) are supervised
%and focus primarily on learning the best combination of numerous features for a non-linear regression
%model. But even the rare, yet quite successful, unsupervised models (Kashyap
%et al., 2014; Sultan et al., 2014) rely on various language-specific tools (e.g.,
%named entity recognition, dependency parsing) and resources, e.g., WordNet
%(Fellbaum, 1998). The fact that such tools and resources exist only for a
%handful of languages also limits the impact of these models.

%sultanovo metoda sice neni supervised STS, ale vyuziva nejruznejsi NERKa, atd...\\

%unsupervised STS - \cite{GLAVAS2017} \cite{sultan:2015} \cite{brychcin-svoboda:2016:SemEval} \\

Lately, research in NLU is moving beyond monolingual meaning comparison. The research is motivated mainly by two factors: a) cross-lingual semantic similarity metrics enable us to work in multilingual contexts, which is useful in many applications (cross-lingual information retrieval, machine translation, etc.) and b) it enables transferring of knowledge between languages, especially from resource-rich to poorly-resourced languages. In the last two years, STS shared tasks \cite{agirre-EtAl:2016:SemEval1,cer-EtAl:2017:SemEval} have been extended by cross-lingual tracks. The best performing systems \cite{brychcin-svoboda:2016:SemEval,tian-EtAl:2017:SemEval} first employ an off-the-shelf machine translation service to translate input sentences into the same language and then apply state-of-the-art monolingual STS models. These highly-tuned approaches rely on manually annotated data, numerous resources, and tools, which significantly limit their applicability for poorly-resourced languages. Unlike the prior works, we study purely unsupervised STS techniques based on word distributional-meaning representations as the only source of information.

The fundamental assumption (\textit{Distributional Hypothesis})
is that two words are expected to be semantically similar if they
occur in similar contexts (they are similarly distributed across
the text). This hypothesis was formulated by \newcite{Harris:1954}
several decades ago. Today it is the basis of state-of-the-art distributional semantic models \cite{DBLP:journals/corr/abs-1301-3781,pennington-socher-manning:2014:EMNLP2014,TACL999}. Unsupervised methods for assembling word %meaning 
representations to estimate textual similarity have been proposed in \cite{brychcin-svoboda:2016:SemEval,mu-bhat-viswanath:2017:Short,GLAVAS2017}. We describe them in detail in Section \ref{sec:STS}.

Several approaches for inducing cross-lingual word semantic representation (i.e., unified semantic space for different languages) have been proposed in recent years, each requiring a different form of cross-lingual supervision \cite{upadhyay-EtAl:2016:P16-1}. They can be roughly divided into three categories according to the level of required alignment: a) document-level alignments \cite{Vulic2016BilingualDW}, b) sentence-level alignments \cite{levy-sogaard-goldberg:2017:EACLlong}, and c) word-level alignments \cite{DBLP:journals/corr/MikolovLS13}. 

We focus on the last case, where a common approach is to train monolingual semantic spaces independently of each other and then to use bilingual dictionaries to transform semantic spaces into a unified space. Most related works rely on linear transformations \cite{DBLP:journals/corr/MikolovLS13,faruqui-dyer:2014:EACL,lazaridou-dinu-baroni:2015:ACL-IJCNLP,artetxe-labaka-agirre:2016:EMNLP2016} and profit from weak supervision. \newcite{vulic-korhonen:2016:P16-1} showed that bilingual dictionaries with a few thousand word pairs are sufficient.  
Such dictionaries can be easily obtained for most languages. Moreover, the mapping between semantic spaces can be easily extended to a multilingual scenario %(more than two languages) 
\cite{DBLP:journals/corr/AmmarMTLDS16}.

This paper investigates linear transformations for cross-lingual STS. We see two contributions of our work:

%\vspace{-0.3cm}
\begin{itemize}
\item We propose a new linear transformation, which outperforms others in the cross-lingual STS task on several datasets.
\item We extend previously published methods for unsupervised STS by word weighting. This leads to significantly better results.
%\item We provide thorough comparison of several linear transformations and several methods for STS.
\end{itemize}
%\vspace{-0.2cm}

This paper is organized as follows. 
In Section \ref{sec:STS}, we start with description of STS techniques based on combining word representations.
The process of learning cross-lingual word representations via linear transformations is explained in Section \ref{sec:transform}.  We propose our transformation in Section \ref{sec:ort}. We show our experimental results in Section \ref{sec:experiments} and conclude in Section \ref{sec:conclusion}.

\section{Semantic Textual Similarity\label{sec:STS}}
%\vspace{-0.2cm}

Let $w \in \boldsymbol V$ denote the word, where $\boldsymbol V$ is a vocabulary. Let $\mathcal{S}: \boldsymbol V \mapsto \mathbb{R}^d$ be a semantic space, i.e., a function which projects word $w$ into Euclidean space with dimension $d$. The meaning of the word $w$ is represented as a real-valued vector $\mathcal{S} (w)$.
We assume \emph{bag-of-words} principle and represent the sentence as a set (bag) $\boldsymbol w = \{w \in \boldsymbol V\}$, i.e., the word order has no role. Note we allow repetitions of the same word in the sentence (set).  Given two sentences $\boldsymbol w^x$ and $\boldsymbol w^y$, the task is to estimate their semantic similarity
$sim(\boldsymbol w^x,\boldsymbol w^y) \in \mathbb{R}$. 

\newcite{brychcin-svoboda:2016:SemEval} showed that inverse-document-frequency (IDF) weighting can boost STS performance. Inspired by their approach, we assume not all words in vocabulary $\boldsymbol V$ are of the same importance. We represent this importance by IDF weight $\lambda_w = idf(w)$. The importance of sentence $\boldsymbol w$ is then represented as a sum of corresponding word weights $ \lambda_{\boldsymbol w} = \sum_{w \in \boldsymbol w} \lambda_w$.
%If the sentences are in different languages we assume separate vocabularies $\boldsymbol V^x$ and $\boldsymbol V^y$, weighting $\lambda^x_{w^x}$ and $\lambda^y_{w^y}$, and semantic spaces $\mathcal{S}^x$ and $\mathcal{S}^y$.

In the following text we describe three STS approaches, which rely only on the word meaning representations, and extend them to incorporate word weights $\lambda_w$. 
For the original version of STS algorithms, we consider uniform weighting of words, i.e., $\lambda_w=1$ for all $w \in \boldsymbol V$.

%\vspace{-0.1cm}
\subsection{Linear Combination}
%\vspace{-0.1cm}

Following \emph{Frege's principle of compositionality} \cite{Pelletier1994}, which states that the meaning of a complex expression is determined as a composition of its parts (i.e., words), we represent the meaning of the sentence as a linear combination of word vectors $\mathbf{v}_{\boldsymbol w} = \frac{\sum_{w \in \boldsymbol w} \lambda_w \mathcal{S}(w) }{ \lambda_{\boldsymbol w} }$. 
%\newcite{brychcin-svoboda:2016:SemEval} showed that this approach leads to very good representation of short sentences. 
The similarity between sentences is then calculated as a cosine of angle between sentence vectors

%\vspace{-0.3cm}
\begin{equation}
sim(\boldsymbol w^x,\boldsymbol w^y) = cos( \mathbf{v}_{\boldsymbol w^x} , \mathbf{v}_{\boldsymbol w^y} ).
\end{equation}

%\vspace{-0.1cm}
\subsection{Principal Angles}
%\vspace{-0.1cm}

\newcite{mu-bhat-viswanath:2017:Short} observed that the most information about words in a sentence is encoded in a low-rank subspace. Consequently, similar sentences should have similar subspaces. Using \emph{Principal Component Analysis}, a technique for dimensionality reduction, we can find the linear subspace with most of the variance in word vectors.

Let $\mathbf{W} \in \mathbb{R}^{d \times |\boldsymbol w|}$ denote the sentence matrix with column vectors given by $\lambda_w \mathcal{S}(w)$ for all $w \in \boldsymbol w$. Using \emph{Singular Value Decomposition} (SVD) we decompose the matrix into $\mathbf{W} = \mathbf{U} \mathbf{\Sigma} {\mathbf{V}}^\top$. The matrix $\mathbf{U}_r$ is obtained by truncating the matrix $\mathbf{U}$ to keep only first $r$ principal components. Finally, the similarity between two sentences $\boldsymbol w^x$ and $\boldsymbol w^y$ is defined as L2-norm of the singular values between corresponding subspaces $\mathbf{U}_r^x$ and $\mathbf{U}_r^y$

%\vspace{-0.3cm}
\begin{equation}
sim(\boldsymbol w^x,\boldsymbol w^y) = \sqrt{\sum_{i=1}^{r}{\sigma_i^2}},
\end{equation}

\noindent
where $\sigma_i$ denote the $i$-th singular value of matrix ${\mathbf{U}_r^x}^\top \mathbf{U}_r^y$. Note the singular values are in fact the cosines of the principal angles and can be obtained by SVD of ${\mathbf{U}_r^x}^\top \mathbf{U}_r^y$ from the diagonal matrix $\mathbf{\Sigma}$.

%According to \newcite{mu-bhat-viswanath:2017:Short}, this method outperforms linear combination of word vectors on several datasets.

%\vspace{-0.1cm}
\subsection{Optimal Matching}
%\vspace{-0.1cm}

The method presented in \cite{sultan:2015} has been very successful in STS tasks in the last years. 
%Given two sentences we want to compare, this method finds and aligns the words that have similar meaning and similar role in these sentences. 
This method finds and aligns the words that have similar meaning and similar role in the input sentences. 
The method is considered to be unsupervised in the sense it does not require sentence similarity judgments, however it relies on various language-specific tools (e.g., named entity recognition, dependency parsing, etc.).

Following this approach, \newcite{GLAVAS2017} introduced the unsupervised word alignment method, which relies only on word meaning representations. Let $ \boldsymbol M \subset \boldsymbol w^x \times \boldsymbol w^y$ denote the matching for input sentences $\boldsymbol w^x$ and $\boldsymbol w^y$ consisting of aligned word pairs $ (w^x, w^y) \in \boldsymbol M$. Every word in both sentences can be at most in one pair. 
%If one sentence is longer than the second one, some words remain unaligned. 
In fact we have a complete bipartite graph with nodes represented by words in input sentences, where the weight for an edge is the cosine similarity $\delta_{w^x,w^y} =  cos\big(\mathcal{S}^x(w^x), \mathcal{S}^y(w^y)\big)$. 
The task is to find an optimal matching (alignment) $\hat{\boldsymbol M}$ with the highest sum of cosine similarities between words in pairs.
%
%The task is to find an optimal matching (alignment) with maximum cosine similarities between words in pairs 
%\vspace{-0.3cm}
%\begin{equation}
%\hat{\boldsymbol M} = \argmax_{\boldsymbol M} \quad \sum_{\quad\mathclap{(w^x, w^y) \in \boldsymbol M}} { \delta_{w^x,w^y} }.
%\end{equation}
%
%\noindent
It can be found by so called \emph{Hungarian Method} \cite{Kuhn55thehungarian}. % (improved version has complexity $\mathcal{O}(n^3)$).
%
%Once we have the optimal matching $\hat{\boldsymbol M}$
Finally, we estimate the matching score for $\boldsymbol w^x$ as $\delta_{\boldsymbol w^x} = \frac{1}{\lambda_{\boldsymbol w^x}}\sum_{(w^x, w^y) \in \hat{\boldsymbol M}} { \lambda_{w^x} \delta_{w^x,w^y}}$ (analogously for $\boldsymbol w^y$). The similarity between sentences $\boldsymbol w^x$ and $\boldsymbol w^y$ is then calculated as an average over both scores

%\vspace{-0.3cm}
\begin{equation}
sim(\boldsymbol w^x,\boldsymbol w^y) =  \frac{\delta_{\boldsymbol w^x} + \delta_{\boldsymbol w^y}}{2}.
\end{equation}

\section{Linear Transformations% between Semantic Spaces
\label{sec:transform}}
%\vspace{-0.2cm}

%Given a set of languages $\boldsymbol L$, let word $w^x \in \boldsymbol V^x$ denote the word in language $x \in \boldsymbol L$, where $\boldsymbol V^x$ is a vocabulary of that language. Let $S^x: \boldsymbol V^x \mapsto \mathbb{R}^d$ be a semantic space for language $x$, i.e., a function which projects the words $w^x$ into Euclidean space with dimension $d$. The meaning of the word $w^x$ is represented as a real-valued vector $S^x (w^x)$. We assume the same dimension $d$ for all languages\footnote{Note that all described linear transformations can be easily extended to the general case, where the dimension of two semantic spaces differs.}.

%This paper focuses on linear transformations between semantic spaces. A linear transformation can be expressed as
A linear transformation between semantic spaces can be expressed as

%\vspace{-0.3cm}
\begin{equation}
 \mathcal{S}^{x\rightarrow y}(w^x) =   \mathbf{T}^{x\rightarrow y} \mathcal{S}^x (w^x),
\end{equation}

\noindent
i.e., as a multiplication by a matrix $\mathbf{T}^{x\rightarrow y} \in \mathbb{R}^{d \times d}$.
Linear transformation can be used to perform \textit{affine transformations} (e.g., rotation, reflection, translation, scaling, etc.) and other transformations (e.g., column permutation) \cite{nomizu1994affine}%
\footnote{In the general case, affine transformation is the composition of two functions (a translation and a linear map) represented as $\mathbf{y} = \mathbf{A}\mathbf{x} + \mathbf{b}$. Using so called \textit{augmented matrix} (which extends the dimension by 1), we can rewrite this to\\ $
\begin{vmatrix}
\mathbf{y} \\ 
1 
\end{vmatrix}
=
\begin{vmatrix}
\mathbf{A} & \mathbf{b}\\ 
0 \dots 0 & 1
\end{vmatrix}
\begin{vmatrix}
\mathbf{x} \\ 
1 
\end{vmatrix}
$, i.e., we can use only matrix multiplication (linear map). In our case, we omit this trick and use only matrix $\mathbf{A}$
similarly to all other prior works on linear transformations for cross-lingual NLU. Moreover, in our experiments (Section \ref{sec:exp}), we center both source and target semantic spaces towards zero so that no translation is required.%
%, assuming that $d-1$ coordinates serve for the meaning representation and the one is for the transformation purposes (note it is not constant thus we are talking about approximation). We assume the impact on the final performance is negligible, because we work with high dimension.
}%
. Composition of such operations is a matrix multiplication, which leads again to a matrix in $\mathbb{R}^{d \times d}$.
For estimating the transformation matrix $\mathbf{T}^{x\rightarrow y}$, we use a bilingual dictionary (set of $n$ word pairs) $(w^x, w^y) \in \boldsymbol D^{x \rightarrow y}$, where $\boldsymbol D^{x\rightarrow y} \subset \boldsymbol V^x \times \boldsymbol V^y$ and $| \boldsymbol D^{x\rightarrow y}| = m$. 
%In our case, we translated the original word forms $w^x$ in language $x$ into language $y$ via \textit{Google translate} (see Section \ref{sec:settings}).
 Finally, we use these $m$ aligned word pairs $(w^x, w^y)$ with their corresponding semantic vectors $(\mathcal{S}^x (w^x), \mathcal{S}^y (w^y))$ to form matrices $\mathbf{X} \in \mathbb{R}^{m \times d}$ and $\mathbf{Y} \in \mathbb{R}^{m \times d}$.

In the following sections, we discuss five approaches (including one we propose) for estimating $\mathbf{T}^{x\rightarrow y}$. The optimal transformation matrix with respect to the corresponding criteria is denoted as $\hat{\mathbf{T}}^{x\rightarrow y}$.

%\vspace{-0.1cm}
\subsection{Least Squares Transformation}
%\vspace{-0.1cm}

Following \newcite{DBLP:journals/corr/MikolovLS13}, we can estimate the matrix $\mathbf{T}^{x\rightarrow y}$ by minimizing the sum of squared residuals. The optimization problem is %given by

%\vspace{-0.5cm}
\begin{equation}
\hat{\mathbf{T}}^{x\rightarrow y} = \argmin_{\mathbf{T}^{x\rightarrow y}} \big\|{\mathbf{Y} - \mathbf{X}  \mathbf{T}^{x\rightarrow y}}\big\|_2^2 
\end{equation}
%\vspace{-0.5cm}

\noindent
and can be solved for example by the gradient descent algorithm. 
The least squares method also has an analytical solution. By taking the Moore-Penrose pseudo-inverse of $\mathbf{X}$, which can be computed using SVD %\cite{doi:10.1137/1.9780898719048}, we achieve
, we achieve

%\vspace{-0.3cm}
\begin{equation}
\hat{\mathbf{T}}^{x\rightarrow y} = (\mathbf{X}^\top \mathbf{X})^{-1}\mathbf{X}^\top \mathbf{Y}.
\end{equation}
%\vspace{-0.7cm}

%\newcite{lazaridou-dinu-baroni:2015:ACL-IJCNLP} showed that the least squares mapping leads to increasing the \textit{hubness} in the final space, because the set of vectors in $\mathbf{X}  \hat{\mathbf{T}}^{x\rightarrow y}$ has lower variance than in $\mathbf{Y}$ (points are on average closer to each other).

%\vspace{-0.2cm}
\subsection{Orthogonal Transformation\label{sec:ot}}
%\vspace{-0.1cm}

%Motivated by inconsistency among the objective functions for learning word representations (based on dot products), the least squares mapping (minimizing Euclidean distances), and word similarity evaluation (based on cosine similarities), \newcite{xing-EtAl:2015:NAACL-HLT} argued that the transformation matrix in the least squares objective should be orthogonal. For estimating this matrix, they introduced an approximate algorithm composed of gradient descent updates and repeated applications of the SVD. \newcite{artetxe-labaka-agirre:2016:EMNLP2016} then derived the analytical solution for the orthogonality constraint and showed that this transformation preserves the monolingual performance of the source space.

\newcite{artetxe-labaka-agirre:2016:EMNLP2016} argued that the transformation matrix in the least squares objective should be orthogonal, because it preserves the angles between points in the space. They derived the analytical solution for the orthogonality constraint and showed that the semantic space has the same monolingual performance after the transformation.

Orthogonal transformation is the least squares transformation subject to the constraint that the matrix $\mathbf{T}^{x\rightarrow y}$ is orthogonal\footnote{Matrix $\mathbf{A}$ is orthogonal if contains orthonormal rows and columns, i.e., $\mathbf{A} \mathbf{A}^\top = \boldsymbol I$. An orthogonal matrix preserves the dot product, i.e., $\mathbf{x} \cdot \mathbf{y} = (\mathbf{Ax}) \cdot (\mathbf{Ay})$, thus the monolingual invariance property.}. The optimal transformation matrix is given by

%\vspace{-0.3cm}
\begin{equation}
\hat{\mathbf{T}}^{x\rightarrow y} = \mathbf{V}\mathbf{U}^\top,
\end{equation}

\noindent
where matrices $\mathbf{V}$ and $\mathbf{U}$ are obtained using SVD of $\mathbf{Y}^\top \mathbf{X}$ \big(i.e., $\mathbf{Y}^\top \mathbf{X} = \mathbf{U} \Sigma \mathbf{V}^\top$\big).

%\vspace{-0.1cm}
\subsection{Canonical Correlation Analysis}
%\vspace{-0.1cm}

Canonical Correlation Analysis is a way of measuring the linear relationship between two multivariate variables (i.e., vectors). It finds basis vectors for each variable in the pair such that the correlation between the projections of the variables onto these basis vectors is mutually maximized.

Given the sample data $\mathbf{X}$ and $\mathbf{Y}$, at the first step we look for a pair of projection vectors $(\mathbf{c}^x_1 \in \mathbb{R}^d, \mathbf{c}^y_1 \in \mathbb{R}^d)$ (also called \textit{canonical directions}), whose data projections $(\mathbf{X} \mathbf{c}^x_1, \mathbf{Y} \mathbf{c}^y_1)$ yield the largest Pearson correlation. Once we have the best pair, we ask for the second-best pair. On either side of $a$ and $b$, we look for $\mathbf{c}^x_2$ and $\mathbf{c}^y_2$ in the subspaces orthogonal to the first canonical directions $\mathbf{c}^x_1$ and  $\mathbf{c}^y_1$, respectively, maximizing correlation of data projections. Generally, $k$-th canonical directions are given by
%
%\vspace{-0.5cm}
%\begin{equation}
$(\mathbf{c}^x_k, \mathbf{c}^y_k) = \argmax_{\mathbf{c}^x, \mathbf{c}^y} {corr (\mathbf{X} \mathbf{c}^x, \mathbf{Y} \mathbf{c}^y)}$,  
%\end{equation}
%\vspace{-0.3cm}
%
%\noindent
where $(\mathbf{X} \mathbf{c}^x) \cdot (\mathbf{X} \mathbf{c}^x_i) = 0$ and $(\mathbf{Y} \mathbf{c}^y) \cdot (\mathbf{Y} \mathbf{c}^y_i) = 0$, for each $1 \le i < k$. In the end of this process, we have bases of $d$ canonical directions for both sides $x$ and $y$. We can represent them as a pair of matrices $\mathbf{C}^x \in \mathbb{R}^{d \times d}$ and $\mathbf{C}^y \in \mathbb{R}^{d \times d}$ (each column corresponds to one canonical direction $\mathbf{c}_k^x$ or $\mathbf{c}_k^y$, respectively), which project $\mathbf{X}$ and $\mathbf{Y}$ into a shared space. The exact algorithm for finding these bases is described in \cite{Hardoon:2004}.

\newcite{faruqui-dyer:2014:EACL} used Canonical Correlation Analysis for incorporating multilingual contexts into word representations, outperforming the standalone monolingual representations on several intrinsic evaluation metrics. \newcite{DBLP:journals/corr/AmmarMTLDS16} extended this work and created a multilingual semantic space for more than fifty languages. Following their approach, the final linear transformation is given by

%\vspace{-0.3cm}
\begin{equation}
\hat{\mathbf{T}}^{x\rightarrow y} = \mathbf{C}^x ({\mathbf{C}^y})^{-1}.
\end{equation}

%\vspace{-0.3cm}
\subsection{Ranking Transformation}
%\vspace{-0.1cm}

All three previous transformations have analytical solutions and employ SVD. \newcite{lazaridou-dinu-baroni:2015:ACL-IJCNLP} studied the \emph{hubness} problem in high-dimensional spaces and point out that cross-lingual transformations should focus on it. They use \textit{max-margin hinge loss} to estimate the transformation matrix and significantly reduce the hubness in the cross-lingual space.

Given the correct word vector $\mathbf{y}$ in target matrix $\mathbf{Y}$ and our estimation $\hat{\mathbf{y}} = \mathbf{x} \mathbf{T}^{x\rightarrow y}$, we want the correct vector $\mathbf{y}$ to be ranked better (to be in front places in the nearest-neighbor list of $\hat{\mathbf{y}}$) than any other word vector $\mathbf{n}$ in $\mathbf{Y}$.  The rank for given pair of vectors $\hat{\mathbf{y}}$ and $\mathbf{y}$  is defined as

%\vspace{-0.6cm}
\begin{equation}
\mathcal{R}(\hat{\mathbf{y}}, \mathbf{y}, \boldsymbol N)=\sum_{\mathbf{n} \in \boldsymbol N} {\big| \gamma + \mathcal{D}(\hat{\mathbf{y}},\mathbf{y})-\mathcal{D}(\hat{\mathbf{y}},\mathbf{n}) \big|_+},
\end{equation}
%\vspace{-0.4cm}

\noindent
where $\boldsymbol N = \mathcal{N}(\hat{\mathbf{y}}, \mathbf{y}, \mathbf{Y})$ is a function, which returns the set of negative examples $\boldsymbol N$, from which we want to get away (in this case containing all word vectors from $\mathbf{Y}$ except $\mathbf{y}$). Let $\mathcal{D}(\hat{\mathbf{y}},\mathbf{y})$ be a distance metric according to which we define the neighborhood.  The ranking function incorporates only those vectors, which were ranked better than $\mathbf{y}$, i.e., $|t|_+ \overset{\scriptscriptstyle def}{=} \max(0, t)$. The margin $\gamma$ means the minimal lead that the correct vector $\mathbf{y}$ should have over other vectors. The lower $\mathcal{R}(\hat{\mathbf{y}}, \mathbf{y}, \boldsymbol N)$ is, the better position in the nearest-neighbors list $\mathbf{y}$ has. The final optimization function is then

%\vspace{-0.6cm}
\begin{equation}
\hat{\mathbf{T}}^{x\rightarrow y} = \argmin_{\mathbf{T}^{x\rightarrow y}} \sum_{i=1}^{m} {\mathcal{R}\big(\hat{\mathbf{y}_i}, \mathbf{y}_i, \mathcal{N}(\hat{\mathbf{y}_i}, \mathbf{y}_i, \mathbf{Y})\big)},
\end{equation}
%\vspace{-0.4cm}

\noindent
where $\mathbf{x}_i$ and $\mathbf{y}_i$ is an $i$-th row vector in matrix $\mathbf{X}$ and $\mathbf{Y}$, respectively. The optimum can be found for example by gradient descent algorithm.

In practice this is computationally too expensive as it requires summation over $m \times (m-1)$ elements. \newcite{lazaridou-dinu-baroni:2015:ACL-IJCNLP} argue this can be solved by using only single negative example 

%\vspace{-0.7cm}
\begin{multline}
\label{eq:negativesamples}
\mathcal{N}(\hat{\mathbf{y}_i}, \mathbf{y}_i, \mathbf{Y}) = \\
\Big\{ \argmin_{1 \le j \le m, j \neq i} \big[\mathcal{D}(\hat{\mathbf{y}_i},\mathbf{n}_j) - \mathcal{D}(\mathbf{y}_i,\mathbf{n}_j) \big] \Big\},
\end{multline}
%\vspace{-0.4cm}

\noindent
which is as near as possible to our estimation $\hat{\mathbf{y}_i}$ and as far as possible from the correct vector $\mathbf{y}_i$, i.e., it represents a serious mistake and thus should be the most informative. However, the loss function is not convex anymore and the gradient descent algorithm may not find an optimal solution.

%\vspace{-0.1cm}
\section{Proposed Transformation\label{sec:ort}}
%\vspace{-0.2cm}
\subsection{Current Issues}
%\vspace{-0.1cm}

\newcite{lazaridou-dinu-baroni:2015:ACL-IJCNLP} mentioned \emph{overfitting} as the problem of current linear transformations for cross-lingual spaces (including one they introduced).
%\newcite{lazaridou-dinu-baroni:2015:ACL-IJCNLP} mentioned that the problem of current linear transformations for cross-lingual spaces (including one they introduced) is that they overfit on bilingual dictionaries. 
Several authors extended their learning objectives with $L_2$ regularization term forcing the values in $\mathbf{T}$ towards zero. \newcite{xing-EtAl:2015:NAACL-HLT} and \newcite{artetxe-labaka-agirre:2016:EMNLP2016} experimented with orthogonality constraints on $\mathbf{T}$ forcing all vectors in $\mathbf{T}$ to be orthonormal.

\newcite{Radovanovic:2010} defined \emph{hubness} as one of the curses of dimensionality. \newcite{lazaridou-dinu-baroni:2015:ACL-IJCNLP} derived that the least squares transformation directly leads to worse hubness in the final space and they showed that their ranking transformation can effectively deal with this issue. In our experiments in Section \ref{sec:hubness}, we support this claim and show that other transformations also suffer from the same issue.

We believe another issue is \emph{asymmetry}. The learning objectives of current cross-lingual transformations focus only on one direction, i.e., to approximate the target matrix $\mathbf{Y}$ with an estimation $\hat{\mathbf{Y}} = \mathbf{X} \mathbf{T}^{x\rightarrow y}$ subject to some optimization criteria. However, this does not necessary mean that $({\mathbf{T}^{x\rightarrow y}})^{-1}$ is optimal for approximation of $\mathbf{X}$ by $\hat{\mathbf{X}} = \mathbf{Y} ({\mathbf{T}^{x\rightarrow y}})^{-1}$ in the same objective.

%\vspace{-0.1cm}
\subsection{Orthogonal Ranking Transformation}
%\vspace{-0.1cm}

In this work we address all three mentioned issues (i.e., overfitting, hubness, and asymmetry) and propose a linear transformation with two max-margin loss functions in a single objective

%\vspace{-0.8cm}
\begin{multline}
\label{eq:ort}
\hat{\mathbf{T}}^{x\rightarrow y} = \argmin_{\mathbf{T}^{x\rightarrow y}}  \\
\sum_{i=1}^{m} \Big[ \mathcal R \big(\hat{\mathbf{y}}_i, \mathbf{y}_i, \mathcal{N}(\hat{\mathbf{y}_i}, \mathbf{y}_i, \mathbf{Y})\big) + \\
\mathcal R \big(\hat{\mathbf{x}}_i, \mathbf{x}_i, \mathcal{N}(\hat{\mathbf{x}_i}, \mathbf{x}_i, \mathbf{X})\big) \Big].
\end{multline}
%\vspace{-0.5cm}

\noindent
The key idea is that $\hat{\mathbf{y}} = \mathbf{x} \mathbf{T}^{x\rightarrow y}$ and simultaneously $\hat{\mathbf{x}} = \mathbf{y} {\mathbf{T}^{x\rightarrow y}}^\top$, which has several positive consequences. The objective forces matrix $\mathbf{T}^{x\rightarrow y}$ to be \emph{nearly orthogonal}, i.e., ${(\mathbf{T}^{x\rightarrow y}})^{-1} \approx {\mathbf{T}^{x\rightarrow y}}^\top$. The term nearly orthogonal means that the orthogonality is not a hard constraint as in \cite{artetxe-labaka-agirre:2016:EMNLP2016}, but a regularization condition, which may be only partially fulfilled by the training. The orthogonality constraint makes the learning process much simpler. Estimating inverse matrix during the gradient descent steps would otherwise be computationally unfeasible. Moreover, orthogonality proved to preserve the monolingual performance of semantic spaces.

Each ranking function $\mathcal R$ optimizes
%Both ranking functions optimize 
the mapping in one direction $x\rightarrow y$ and $y \rightarrow x$, which means that our objective is independent of the mapping direction. Consequently, $\hat{\mathbf{T}}^{y\rightarrow x} \approx {(\hat{\mathbf{T}}^{x\rightarrow y}})^\top$.

The two ranking functions address the hubness and rank the correctly mapped vectors in both directions higher than the others. This leads to the lowest hubness between semantic spaces, compared with other transformations (see Section \ref{sec:hubness}).

%\vspace{-0.2cm}
\section{Experiments\label{sec:experiments}}
%\vspace{-0.2cm}
\subsection{Settings\label{sec:settings}}
%\vspace{-0.1cm}

We experiment with all five techniques for linear mapping (all described in sections \ref{sec:transform} and \ref{sec:ort}), namely, \emph{Least Squares Transformation} (LS), \emph{Orthogonal Transformation} (OT), \emph{Canonical Correlation Analysis} (CCA), \emph{Ranking Transformation} (RT), and proposed \emph{Orthogonal Ranking Transformation} (ORT). We combine word representations to estimate semantic textual similarity by all three methods described in Section \ref{sec:STS}, i.e., \emph{Linear Combination} (LC), \emph{Principal Angles} (PA), and \emph{Optimal Matching} (OM). In case of PA, we use $r=4$ principal components as recommended by \newcite{mu-bhat-viswanath:2017:Short}. We evaluate both uniform weighting (for mutual comparison with original methods) and IDF weighting in all three STS approaches.

Our experiments start with building monolingual semantic
spaces for each of tested languages, namely, English (\textsc{En}), Spanish (\textsc{Es}),
Italian (\textsc{It}), Croatian (\textsc{Hr}), Turkish (\textsc{Tr}), and Arabic (\textsc{Ar}). We use character-n-gram-based
skip-gram model \cite{TACL999}, which recently achieved the
state-of-the-art performance in several word similarity and word analogy
tasks for several languages. For all languages except Croatian,
we use word vectors pre-trained on Wikipedia\footnote{%
%Semantic spaces for many languages trained on Wikipedia are available to download at \url{https://fasttext.cc}.}. 
Available at \url{https://fasttext.cc}.}. 
The Wikipedia corpus for Croatian yields poor performance, so we
combine it with web-crawled texts. We adopted the corpus
hrWaC\footnote{Available at \url{http://takelab.fer.hr/data}.}
 \cite{vsnajder-pado-agic:2013:Short} and merged it with
Croatian Wikipedia. The final Croatian corpus has approximately 1.3 billion tokens. 
We use settings recommended by \newcite{TACL999},
i.e., texts are lowercased, vector dimension is set to $d=300$,
and character n-grams from 3 to 6 characters are used. 
We keep the vocabulary of $\boldsymbol V=300$k most frequent words for all languages.
We estimate IDF weights on the Wikipedia corpus for every language. Each Wikipedia article represents a document.

Bilingual dictionaries $\boldsymbol D^{x \rightarrow y}$ between each pair of languages $x$ and $y$, are created from the $m$ most
frequent words in corpus of language $x$ and their translation into language $y$ using Google translate.

We experimented with different global post-processing techniques for semantic spaces. The best setting consists of two steps. Firstly, we move the space towards zero (column-wise mean centering), which is a standard step in regression analysis. \newcite{artetxe-labaka-agirre:2016:EMNLP2016} showed this improves results of linear mappings. Secondly, we normalize word vectors to be unit vectors. This guarantees that all word pairs in dictionary $\boldsymbol D^{x \rightarrow y}$ contribute equally to the optimization criteria of linear transformation. We always apply this post-processing for both semantic spaces $\mathcal{S}^x$ and $\mathcal{S}^y$ in a pair before the linear mapping.

RT and ORT require special settings to work properly. Unlike the original ranking transformation \cite{lazaridou-dinu-baroni:2015:ACL-IJCNLP}, we do not start the learning process from random values in matrix $\mathbf{T}^{x\rightarrow y}$, but we initialize it by the orthogonal transformation (Section \ref{sec:ot}). We use 50 negative samples minimizing Equation \ref{eq:negativesamples} (in case of ORT, we use 50 negative samples for each side). We set $\gamma$ to 0. \newcite{lazaridou-dinu-baroni:2015:ACL-IJCNLP} used inverse cosine as a distance metric $\mathcal{D}(\hat{\mathbf{y}},\mathbf{y})$. However, in our experiments Euclidean distance yields slightly better results so we choose this as a final distance metric. Five training epochs (more information in Section \ref{sec:learning}) was enough for estimating parameters for both RT and ORT.

%\vspace{-0.2cm}
\subsection{Evaluation}
%\vspace{-0.1cm}

\begin{figure*}
     \begin{center}
	\begin{subfigure}{0.33\textwidth}
           	\includegraphics[width=\textwidth]{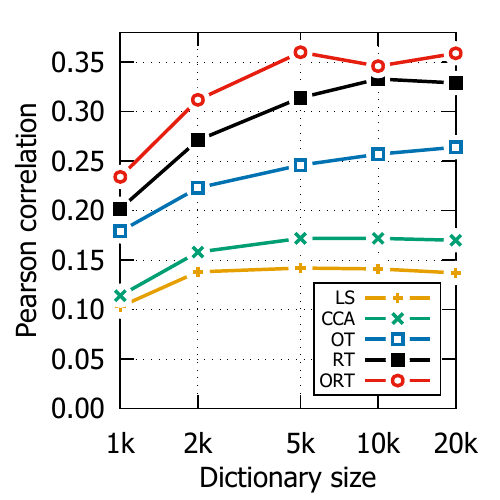}\caption{LC}
      	 \end{subfigure}%
	\begin{subfigure}{0.33\textwidth}
            	\includegraphics[width=\textwidth]{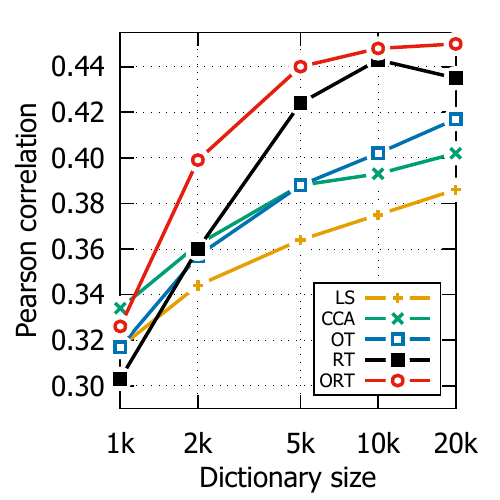}\caption{PA}
        \end{subfigure}%
	\begin{subfigure}{0.33\textwidth}
            	\includegraphics[width=\textwidth]{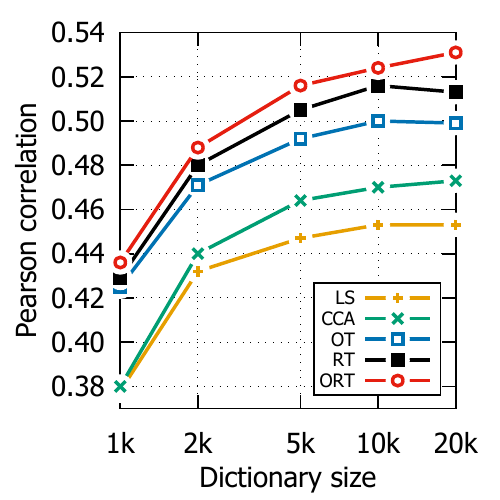}\caption{OM}
        \end{subfigure}
%\vspace{-0.3cm}
\caption{\label{fig:data-size} Ranging dictionary size for all STS techniques with IDF weighting.}
\end{center}
%\vspace{-0.3cm}
\end{figure*}

%STS task has a long history at SemEval workshops and many datasets have been proposed. Evaluation data consist of pairs of sentences ($\boldsymbol w^x$ and $\boldsymbol w^y$) and the degree of their semantic similarity. Performance is measured by the Pearson correlation between automatically estimated scores $sim(\boldsymbol w^x,\boldsymbol w^y)$ and human judgments.

We use datasets from SemEval 2017 task 1 \cite{cer-EtAl:2017:SemEval} as the main evaluation data. They introduced four cross-lingual datasets (\emph{Track4a} and \emph{Track4b} between Spanish and English, \emph{Track2} between Arabic and English, and \emph{Track6} between Turkish and English) and three monolingual datasets (\emph{Track5} in English, \emph{Track3} in Spanish, and \emph{Track1} in Arabic). All datasets contain 250 sentence pairs. In addition, we experiment with data from previous SemEval years. \newcite{agirre-EtAl:2016:SemEval1} proposed two Spanish-English datasets in SemEval 2016 task 1: \emph{News} (301 pairs) and \emph{Multi-source} (294 pairs). \newcite{GLAVAS2017} translated Spanish sentences from both datasets to Italian and Croatian. They also took two English monolingual datasets from SemEval 2012 task 6 \cite{agirre-EtAl:2012:STARSEM-SEMEVAL}, \emph{MSRvid} (750 pairs) and \emph{OnWN} (750 pairs), and translated one side of sentences to Spanish, Italian, and Croatian.

Performance is measured by the Pearson correlation between automatically estimated scores $sim(\boldsymbol w^x,\boldsymbol w^y)$ and human judgments.

%\vspace{-0.2cm}
\subsection{Results\label{sec:exp}}
%\vspace{-0.1cm}

\begin{table}
\setlength{\tabcolsep}{0.2em}
\begin{center}
\resizebox{0.48\textwidth}{!}{
\begin{tabular}{cr|cccccc}
\specialrule{.15em}{.0em}{.0em} 
 & & \bf LC  & \bf LC IDF & \bf PA & \bf PA IDF & \bf OM & \bf OM IDF  \\
\hline
\multicolumn{2}{c|}{Monoling.} &	0.594	&	0.703	&	0.709	&	0.726	&	0.713	&	\bf0.756 \\
\hline
\multirow{5}{*}{\rotatebox[origin=c]{90}{Cross-lingual}} 
& LS			&	0.038	&	0.137	&	0.275	&	0.386	&	0.377	&	\bf0.453 \\
& CCA			&	0.048	&	0.170	&	0.285	&	0.402	&	0.406	&	\bf0.473 \\
& OT			&	0.102	&	0.264	&	0.328	&	0.417	&	0.443	&	\bf0.499 \\
& RT			&	0.191	&	0.329	&	0.374	&	0.435	&	0.468	&	\bf0.513 \\
& ORT			&	0.237	&	0.359	&	0.402	&	0.450	&	0.495	&	\bf0.531 \\
\specialrule{.15em}{.0em}{.0em} 
\end{tabular}
}
\end{center}
%\vspace{-0.4cm}
\caption{The mean Pearson correlations over monolingual and cross-lingual tracks of \mbox{SemEval} 2017 STS dataset.}
\label{tab:global}
%\vspace{-0.5cm}
\end{table}

\begin{table*}[ht!]
\begin{center}
\resizebox{\textwidth}{!}{
\begin{tabular}{r|ccc|c|cccc|c|c}
\specialrule{.15em}{.0em}{.0em} 
& \multicolumn{4}{c|}{\bf Monolingual} &  \multicolumn{5}{c|}{\bf Cross-lingual} & \bf Mean \\
   & \bf \textsc{En}-\textsc{En} & \bf \textsc{Es}-\textsc{Es} & \bf \textsc{Ar}-\textsc{Ar} & & \multicolumn{2}{c}{\bf \textsc{Es}-\textsc{En}}	& \bf \textsc{Ar}-\textsc{En}	 & \bf \textsc{Tr}-\textsc{En} && \bf over \\
    & \bf Track5 & \bf Track3 & \bf Track1 & \bf Mean & \bf Track4a	 & \bf Track4b	 & \bf Track2	 & \bf Track6 	& \bf Mean & \bf all \\
\hline
LS		&	0.786	&	0.792	&	0.690	&	0.756	&	0.624	&	\bf 0.231	&	0.465	&	0.490		&	0.453	&	0.583 \\
CCA		&	0.786	&	0.792	&	0.690	&	0.756	&	0.643	&	0.215	&	0.511	&	0.523		&	0.473	&	0.594 \\
OT		&	0.786	&	0.792	&	0.690	&	0.756	&	0.652	&	0.214	&	0.572	&	0.560		&	0.499	&	0.609 \\
RT		&	0.786	&	0.792	&	0.690	&	0.756	&	0.665	&	0.228	&	0.582	&	0.575		&	0.513	&	0.617 \\
ORT		&	0.786	&	0.792	&	0.690	&	0.756	&\bf 0.685	&	0.219	&\bf 0.618	&	\bf 0.604	&	\bf 0.531	&	\bf 0.628 \\
\hline
\newcite{tian-EtAl:2017:SemEval}	&	0.852	&	0.856	&	0.744	&	0.817	&	0.813	&	0.336	&	0.749	&	0.771	&	0.667	&	0.732 \\
\newcite{cer-EtAl:2017:SemEval}	&	0.728	&	0.712	&	0.605	&	0.682 &	0.622	&	0.032	&	0.516	&	0.546	&	0.429	&	0.537 \\
\specialrule{.15em}{.0em}{.0em} 
\end{tabular}
}
\end{center}
%\vspace{-0.4cm}
\caption{Individual Pearson correlations for all tracks in SemEval 2017 using OM with IDF weighting.}
\label{tab:ind-semeval2017}
%\vspace{-0.1cm}
\end{table*}

\begin{table*}[ht!]
\setlength{\tabcolsep}{0.2em}
\begin{center}
\resizebox{\textwidth}{!}{
\begin{tabular}{r|cccc|cccc|cccc|c}
\specialrule{.15em}{.0em}{.0em} 
 & \multicolumn{4}{c|}{\bf \textsc{It}-\textsc{En}}  & \multicolumn{4}{c|}{\bf \textsc{Hr}-\textsc{En}}  & \multicolumn{4}{c|}{\bf \textsc{Es}-\textsc{En}} & \bf Mean\\
  & \bf \multirow{2}{*}{MSRvid} & \bf \multirow{2}{*}{OnWN} & \bf \multirow{2}{*}{News} & \bf Multi- & \bf \multirow{2}{*}{MSRvid} & \bf \multirow{2}{*}{OnWN} & \bf \multirow{2}{*}{News} & \bf Multi- & \bf \multirow{2}{*}{MSRvid} & \bf \multirow{2}{*}{OnWN} & \bf \multirow{2}{*}{News} & \bf Multi-  & \bf over  \\
  &  & & & \bf source &  &  &  & \bf source &  &  &  & \bf source &  \bf all\\
\hline
LS		&	0.695	&	0.594	&	0.875	&	0.663	&	0.656	&	0.509	&	0.830	&	0.569	&	0.660	&	0.593	&	0.880	&	0.690	&	0.685 \\
CCA		&	0.704	&	0.598	&	0.890	&	0.693	&	0.664	&	0.523	&	0.853	&	0.618	&	0.646	&	0.597	&	0.893	&	0.705	&	0.699 \\
OT		&	0.704	&	0.594	&	0.895	&	0.718	&	0.686	&	0.535	&	0.876	&	0.701	&	0.638	&	0.595	&	0.896	&	0.731	&	0.714 \\
RT		&	0.744	&	0.596	&\bf 0.896		&\bf0.796		&	0.689	&	0.544	&	0.877	&	0.725	&	0.667	&	0.594	&	0.896	&	0.768	&	0.733 \\
ORT		&\bf0.746		&	\bf 0.605	&\bf 0.896		&	0.794	&\bf0.719		&\bf0.546		&\bf0.882		&\bf0.745		&\bf0.697		&	\bf0.602	&\bf0.900		&\bf0.778		&\bf0.743 \\
\hline
\newcite{GLAVAS2017}				&	0.602	&	0.452	&	0.848	&	0.704	&	0.528	&	0.390	&	0.784	&	0.648	&	0.613	&	0.496	&	0.866	&	0.772	&	0.642 \\
\newcite{brychcin-svoboda:2016:SemEval}	&		&		&		&		&		&		&		&		&		&		&	0.906	&	0.819 \\
\specialrule{.15em}{.0em}{.0em} 
\end{tabular}
}
\end{center}
%\vspace{-0.4cm}
\caption{Individual Pearson correlations for translated datasets from SemEval 2012 and 2016 using OM with IDF weighting.}
\label{tab:ind-gg}
%\vspace{-0.4cm}
\end{table*}

Table \ref{tab:global} shows the mean Pearson correlations for individual linear transformations and different STS techniques on cross-lingual STS datasets from SemEval 2017 (i.e., Track4a, Track4b, Track2, and Track6). The size of bilingual dictionaries was set to $m=20$k. For comparison, we also show the mean Pearson correlation over monolingual tracks (Track5, Track3, and Track1) achieved by the same STS techniques. We can see the clear trend in up-to-down and left-to-right direction. ORT significantly outperformed other transformations independently of STS technique. IDF weighting boosts the correlations in all cases and together with OM approach to STS yields the best performance. An interesting fact is that there is a really significant performance drop when LC is used in cross-lingual environment. PA and OM seem to be much more resistant to weaknesses of linear transformations.

How the size of bilingual dictionary affects the mean correlation on the same data can be seen in Figure \ref{fig:data-size}. We show IDF versions of STS techniques as they perform better. All three sub-figures support our choice to use bilingual vocabularies with 20k word pairs.

\begin{figure*}
     \begin{center}
	\begin{subfigure}{0.33\textwidth}
           	\includegraphics[width=\textwidth]{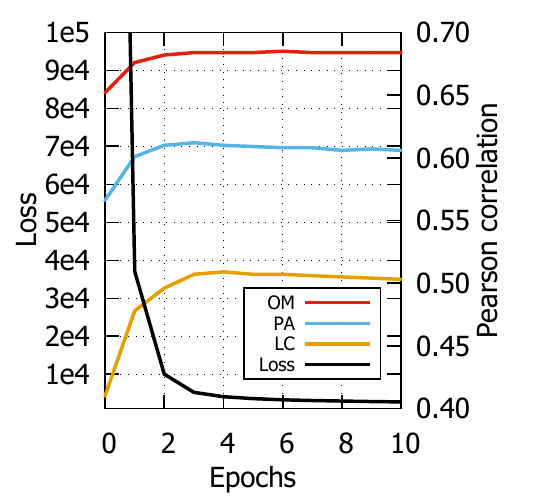}\caption{Track4a (\textsc{Es}$\rightarrow$\textsc{En})}
      	 \end{subfigure}%\hspace{-0.4cm}%
	\begin{subfigure}{0.33\textwidth}
           	\includegraphics[width=\textwidth]{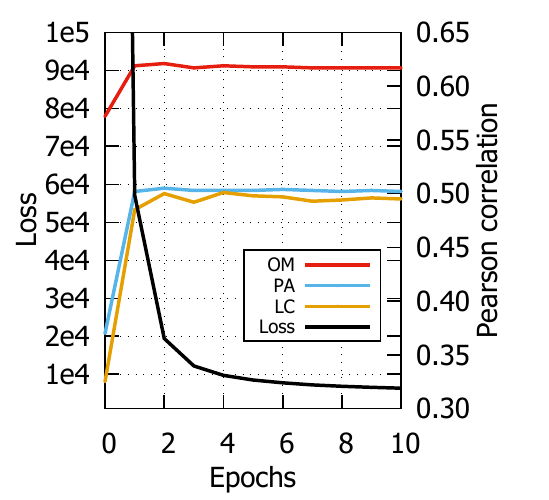}\caption{Track2 (\textsc{Ar}$\rightarrow$\textsc{En})}
      	 \end{subfigure}%\hspace{-0.4cm}%
	\begin{subfigure}{0.33\textwidth}
           	\includegraphics[width=\textwidth]{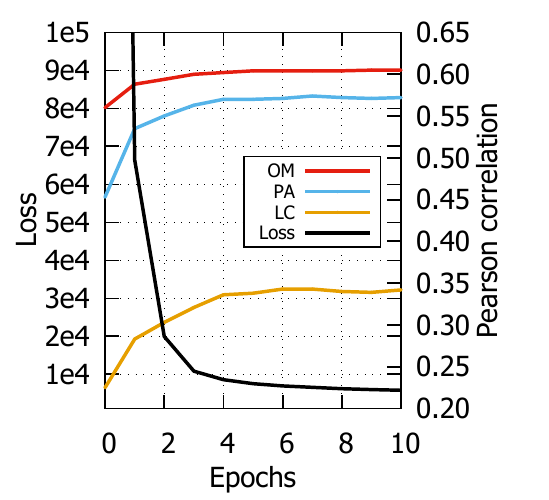}\caption{Track6 (\textsc{Tr}$\rightarrow$\textsc{En})}
      	 \end{subfigure}
%\vspace{-0.3cm}
\caption{ORT learning curves for STS techniques with IDF weighting. Left axis shows the value of loss function, while the right axis shows Pearson correlation on SemEval 2017 cross-lingual tracks.}
\label{fig:learning}
%\vspace{-0.4cm}
\end{center}
\end{figure*}

In tables \ref{tab:ind-semeval2017} and \ref{tab:ind-gg} we show individual correlations achieved by the best settings, i.e., OM with IDF weighting and $m=20$k. Results in Table \ref{tab:ind-semeval2017} are calculated on monolingual and cross-lingual tracks of STS task at SemEval 2017. We compare our results with the top performing system ECNU \cite{tian-EtAl:2017:SemEval}. ECNU employs supervised machine learning with several regression methods and various features together with several deep learning algorithms. They translate all sentences in cross-lingual tracks  into English via Google translate. They achieved mean Pearson correlation 0.817 and 0.667 on monolingual and cross-lingual tracks, respectively. We achieved correlations 0.756 and 0.531 by OM with IDF weighting and bilingual semantic spaces based on ORT transformation. Our approach does not need STS supervision and the only cross-lingual information is a bilingual dictionary. We also compare our results with SemEval baseline \cite{cer-EtAl:2017:SemEval}, i.e., non-English sentences translated into English via Google translate, bag-of-words representation of sentences and cosine similarity. Baseline achieved correlation 0.605 and 0.429 on monolingual and cross-lingual tracks, respectively.

Table \ref{tab:ind-gg} shows results for SemEval 2012 and 2016 datasets translated by \cite{GLAVAS2017}. Again, the best correlations are achieved by weighed OM with ORT transformation (mean Pearson correlation is 0.740). We compare our results with \newcite{GLAVAS2017}, the authors of OM technique. Their system uses LS transformation and unweighed OM, achieving mean Pearson correlation 0.642.  In addition, we show results of UWB system introduced by \newcite{brychcin-svoboda:2016:SemEval}, which performed best on cross-lingual track at SemEval 2016 (i.e., on datasets \emph{News} and \emph{Multi-source}). It uses very similar architecture to ECNU system, i.e., a regression combining state-of-the-art STS methods and features. We can state our results are very competitive.

%\vspace{-0.2cm}
\subsection{Learning\label{sec:learning}}
%\vspace{-0.1cm}

Figure \ref{fig:learning} shows the learning curves for ORT on cross-lingual SemEval 2017 datasets. 
We use stochastic gradient descent algorithm to search the optimal transformation matrix $\hat{\mathbf{T}}$. 
On left vertical axis we see the value of the ORT loss function (i.e., right side of the Equation \ref{eq:ort}). 
Right vertical axis represent Pearson correlations achieved by the weighed versions of STS methods.
The trend is very similar for all three datasets. ORT needs five training epochs to produce optimal results. This setting was used in all our experiments.

%\vspace{-0.2cm}
\subsection{Hubness\label{sec:hubness}}
%\vspace{-0.2cm}

\newcite{Radovanovic:2010} introduced the \emph{hubness} as a new aspect of the dimensionality curse. High-dimensional spaces often contain \emph{hubs}, i.e., points that are near to many other points in the space without being similar in a meaningful way. This affects the distribution of $k$-occurrences very often used in NLU area, e.g., for word analogies \cite{DBLP:journals/corr/abs-1301-3781}, word similarities \cite{camachocollados-EtAl:2017:SemEval}, etc.

Let $N_k(w)$ be a hubness score of $w$, i.e., a number of distinct words for which $w$ occur in the list of $k$-nearest neighbors. \newcite{Radovanovic:2010} showed that $N_k$ distribution becomes considerably skewed as dimensionality increases. For that purpose we can calculate \emph{skewness}, a measure of asymmetry. The skewness of a random variable $N_k$ is the third standardized moment defined as $S_{N_k} = \frac{E(N_k - \mu_{N_k})^3}{\sigma_{N_k}^3 }$, where $ \mu_{N_k}$ and $\sigma_{N_k}$ is the mean and the standard deviation of $N_k$, respectively. The higher skewness means the higher hubness in semantic space.

Table \ref{tab:skewness} shows skewness of semantic spaces used for SemEval 2017 data. Euclidean distance has been used for searching nearest neighbors. Original semantic spaces before transformation have following skewness: English 3.14, Spanish 2.71, Arabic 3.65, and Turkish 4.02, respectively. The columns with an arrow are for cross-lingual skewness, i.e., for words in source semantic spaces we look for nearest neighbors in the target semantic spaces. Columns without arrow denote the skewness in the space after it is transformed onto English. The hubness in the source space remains the same after the transformation by OT, because it preserves the angles between points. Our nearly orthogonal transformation has slightly worse hubness. Surprisingly, not LS, but CCA transformation leads to the worst hubness in the source space.
However, the hubness problem between the semantic spaces is much more important as it affects the cross-lingual performance. In all cases our ORT transformation leads to the lowest hubness. We believe it is because the objective function focused on mapping in both directions.

\begin{table}
\begin{center}
\resizebox{0.48\textwidth}{!}{
\begin{tabular}{r|cc|cc|cc}
\specialrule{.15em}{.0em}{.0em}  
% & \multicolumn{2}{c|}{\bf \textsc{Es}} & \multicolumn{2}{c|}{\bf \textsc{Ar}} & \multicolumn{2}{c}{\bf \textsc{Tr}} \\
  & \bf \textsc{Es} & \bf \textsc{Es}$\rightarrow$\textsc{En}    & \bf \textsc{Ar} & \bf \textsc{Ar}$\rightarrow$\textsc{En}    & \bf \textsc{Tr} & \bf \textsc{Tr}$\rightarrow$\textsc{En}\\
\hline
LS	 	&	3.87	&	6.91	&	8.58	&	11.67&	5.55	&	12.75	\\
CCA 		&	3.47	&	5.69	&	12.97&	6.25	&	6.55	&	10.28	\\
OT 		&\bf2.71	&	5.61	&\bf3.65	&	5.16	&\bf4.02	&	9.18	 \\
RT		&	3.53	&	4.88	&	5.43	&	3.29	&	6.50	&	3.49	 \\
ORT	 	&	3.27	&\bf4.50	&	4.57	&\bf3.18	&	5.31	&\bf3.13	 \\
\specialrule{.15em}{.0em}{.0em} 
\end{tabular}
}
\end{center}
%\vspace{-0.4cm}
\caption{Skewness of hubness distribution $N_{20}$.}
\label{tab:skewness}
%100000 vocabulary, 20 nearest neighbors, en=3,14 es=2,71 ar=3,65 tr=4,02}
%\vspace{-0.5cm}
\end{table}

%\vspace{-0.1cm}
\section{Conclusion and Future Work\label{sec:conclusion}}
%\vspace{-0.2cm}

In this paper we investigated linear transformations to create cross-lingual semantic spaces.  
We introduced a new transformation, which reduces the hubness in semantic spaces.
We used three (previously published) approaches to combine information from word representations. We showed all three approaches can be rapidly improved by a word weighting.
Our STS system does not require sentence similarity supervision and the only cross-lingual information is a bilingual dictionary.

We evaluated on several STS datasets in different languages from SemEval shared tasks.
We showed that with exactly same number of training parameters our transformation yields significantly better performance.
Our unsupervised system provides competitive results even if we compare it with the winning systems from SemEval requiring strong supervision.

As a main direction for future work we plan to evaluate our cross-lingual transformation in different extrinsic tasks (e.g., document classification and syntactic parsing) and intrinsic tasks (e.g., word similarity and word analogy tasks).

\subsubsection*{Acknowledgments.}

This publication was supported by the project LO1506 of the Czech Ministry of Education, Youth and Sports under the program NPU I.

\bibliography{emnlp2018}
\bibliographystyle{acl_natbib}

\end{document}